\pdfoutput=1

\documentclass[11pt]{article}
\usepackage[dvipsnames]{xcolor}

\usepackage{amssymb}
\usepackage[]{acl}
\usepackage{times}
\usepackage{latexsym}

\usepackage{graphicx} 
\usepackage{amsmath}
\usepackage[ruled]{algorithm2e}
\usepackage{multirow}
\usepackage{color}
\usepackage{makecell}
\usepackage{amsmath}
\usepackage{multirow}
\usepackage{makecell}
\usepackage{colortbl}
\definecolor{Gray}{gray}{0.85}
\definecolor{aliceblue}{rgb}{0.94, 0.97, 1.0}
	\definecolor{beaublue}{rgb}{0.74, 0.83, 0.9}
\definecolor{blond}{rgb}{0.98, 0.94, 0.75}
\definecolor{beige}{rgb}{0.96, 0.96, 0.86}
	\definecolor{cornsilk}{rgb}{1.0, 0.97, 0.86}
	\definecolor{platinum}{rgb}{0.9, 0.89, 0.89}
\definecolor{lavendermist}{rgb}{0.9, 0.9, 0.98}
\definecolor{oldlace}{rgb}{0.99, 0.96, 0.9}

\newcommand{\namemodel}{{\textsc{Declare}}} 
\newcommand{\namedata}{{\texttt{MedLane}}} 

\usepackage{times}
\usepackage{latexsym}

\usepackage[T1]{fontenc}

\usepackage[utf8]{inputenc}

\usepackage{microtype}

\title{Benchmarking Automated Clinical Language {Simplification}:\\ Dataset, Algorithm, and Evaluation}

\author{
  Junyu Luo$^1$, Junxian Lin$^{2,3}$, Chi Lin$^{2,3}$, Cao Xiao$^4$, Xinning Gui$^1$, Fenglong Ma$^1${\thanks{\; Corresponding author.}}\\
  $^1$College of Information Sciences and Technology, Pennsylvania State University, USA\\
  $^2$School of Software Technology, Dalian University of Technology, China\\
  $^3$Key Laboratory for Ubiquitous Network and Service Software of Liaoning Province, China\\
  $^4$Relativity, USA\\
  \texttt{\{junyu,xinninggui,fenglong\}@psu.edu,linjunxian@mail.dlut.edu.cn,}\\
  \texttt{c.lin@dlut.edu.cn,cao.xiao@relativity.com}
}

\begin{document}
\maketitle
\begin{abstract}
Patients with low health literacy usually have difficulty understanding medical jargon and the complex structure of professional medical language. Although some studies are proposed to automatically translate expert language into layperson-understandable language, only a few of them focus on both accuracy and readability aspects simultaneously in the clinical domain. 
Thus, simplification of the clinical language is still a challenging task, but unfortunately, it is not yet fully addressed in previous work. To benchmark this task, we construct a new dataset named {\namedata} to support the development and evaluation of automated clinical language simplification approaches. Besides, we propose a new model called {\namemodel} that follows the human annotation procedure and achieves state-of-the-art performance compared with eight strong baselines. To fairly evaluate the performance, we also propose three specific evaluation metrics. Experimental results demonstrate the utility of the annotated {\namedata} dataset and the effectiveness of the proposed model {\namemodel}\footnote{The source code of {\namemodel} and the {\namedata} dataset can be found in the \url{https://github.com/machinelearning4health/MedLane}.}
\end{abstract}

\section{Introduction}
\begin{table*}[]
\centering
\resizebox{\linewidth}{!}{
\begin{tabular}{c|l|c||c|c|c||c|c}
\hline
\multirow{2}{*}{\textbf{\makecell[c]{Text Source}}}        & \multirow{2}{*}{\textbf{\makecell[c]{Dataset Name}}}  & \multirow{2}{*}{\textbf{\makecell[c]{Accessible}}}& \multicolumn{3}{c||}{Term Normalization}                          & \multicolumn{2}{c}{Sentence Simplification}\\ \cline{4-8} 
& & &\multicolumn{1}{c|}{Abbreviations} & \multicolumn{1}{c|}{Acronyms} & \makecell[c]{Complex Phrase}            & \multicolumn{1}{c|}{\makecell[c]{Style Transfer}} & \multicolumn{1}{c}{\makecell[c]{Training Annotations}} \\ \hline

\multirow{4}{*}{\makecell[c]{Biomedical\\Article}} & CLEF~\cite{elhadad2013share}     &\checkmark                      & \checkmark         & \checkmark     & -                  & -                                   & -                                                \\ 
                                    & MSD~\cite{cao2020expertise}     &\checkmark                      & -                                 & -                             & -                         & \checkmark           & -                                         \\ 
                                    & CDSR~\cite{guo2020automated}    &\checkmark                       & -                                 & -                             & -                         & \checkmark           & \checkmark                        \\ 
                                    & \cite{devaraj2021paragraph}    &\checkmark                  & -                                 & -                             & -                         & \checkmark           & \checkmark                        \\ \hline
Perscription                        & e-perscription~\cite{zheng2021work}       &\checkmark          & -                                 & -                             & -                         & \checkmark           & \checkmark                        \\ \hline
\multirow{2}{*}{Wiki}                         & \cite{van2019evaluating}    &-            & -                                 & -                             & -                         & \checkmark           & \checkmark                        \\ 
& AutoMeTS~\cite{van2020automets}     &\checkmark             & -                                 & -                             & -                         & \checkmark           & \checkmark                        \\ \hline
\multirow{3}{*}{\makecell[c]{Clinical\\Note}}                 
                                    & n2c2-track3~\cite{henry20202019}      &\checkmark              & \checkmark         & \checkmark     & -                  & -                                   & -                                                \\ 
                                    & \cite{sakakini2020context}     &-                & -                                 & -                             & -                         & \checkmark           & \checkmark                        \\ 
                                   &  {{\namedata} (Ours) }    \cellcolor{lavendermist}   &\checkmark   \cellcolor{lavendermist}            & \checkmark \cellcolor{lavendermist}        & \checkmark   \cellcolor{lavendermist}  & \checkmark \cellcolor{lavendermist}& \checkmark \cellcolor{lavendermist}          & \checkmark\cellcolor{lavendermist}      \\ \hline      
\end{tabular}
}
\vspace{-0.1in}
\caption{Dataset comparison.}
\label{Tab:CompData}
\vspace{-0.2in}
\end{table*}
Health literacy is generally defined as the ability of patients to obtain, process, understand, and communicate basic health information~\cite{parker1999health}, which is significantly important for making health decisions and ensuring treatment outcomes. The increasing accessibility of technology information makes patients have more opportunities to access health information. However, it is challenging for patients, especially with limited health literacy, to read and understand health materials such as clinical notes written by doctors, with medical jargon~\cite{korsch1968gaps}, abbreviations, and professional language~\cite{friedman2002two}. The lack of proper communication between doctors and patients not only results in a tense doctor-patient relationship~\cite{ha2010doctor} but also increases the risk of adverse health outcomes over time~\cite{sudore2006limited}. Therefore, there is a great need to \emph{simplify professional clinical language to layperson-understandable language}.

\subsection{Why We Need a New Dataset?}~\label{sec:need_data}
To implement an automated clinical language simplification system, the first step is to prepare the dataset for model training. Although there are several annotated medical datasets that are summarized in Table~\ref{Tab:CompData}, most of them do not focus on the clinical domain. As we discussed before, the writing style of clinical materials is significantly different from that used in the general biomedical area. Thus, those datasets cannot be used to train a clinical simplification model. 

To the best of our knowledge, there are only two datasets focusing on clinical language simplification tasks.
The n2c2-track3 dataset~\cite{henry20202019} focuses on clinical term normalization, i.e., recovering the full-term expressions for those abbreviations and acronyms in clinical notes. This term-level recovery cannot guarantee the simplicity of the translated sentences because the full expressions may still be hard to be understood by patients with low levels of health literacy. For example, the full expression of the acronym ``\emph{NC}'' is ``\emph{nasal cannula}'', which is still a professional medical term instead of plain language. 

In~\cite{sakakini2020context}, the authors annotate a clinical language simplification dataset, but it is \textbf{private}. Besides, \emph{the number of the annotated free-text parallel sentences/instances is only 1,541}, which is too small to be enough for evaluating the real performance of machine learning models, especially for deep learning-based models. Finally, this dataset only focuses on sentence-level simplification and \emph{does not provide term-level annotation}, which leads to the difficulty of evaluating whether professional medical terms can be correctly translated and further decreases the reliability of machine learning models.   
Thus, it is essential to create a \textbf{publicly available, large-scale yet fully-annotated} dataset for the clinical language simplification task.

\subsection{Why We Need a New Model?}
Existing models for automatic text simplification (ATS) in the biomedical domain are mainly designed based on the available datasets, which either mainly focus on term normalization or directly apply neural machine translation techniques. 

The term normalization technique aims to recover the full-term expressions for medical abbreviations and acronyms using a dictionary~\cite{2014Mining, elhadad2007mining,deleger2008paraphrase,qenam2017text,rahimi2020wikiumls,liu2021deep}, i.e., only targeting the \emph{term-level simplification} and without considering the readability and understandability of the whole sentences.
Another line of work treats the original sentences as the source language and the simplified sentences as the target language, i.e., the \emph{sentence-level simplification}. They usually borrow the ideas from neural machine translation models to only learn the style mapping function between the original and simplified sentences~\cite{weng2019unsupervised,pattisapu2020leveraging,li2020pharmmt,devaraj2021paragraph} but ignore the term-level translation.

In fact, the drawbacks of existing studies make them impossible to be applied to the new annotated dataset. Therefore, we need to design a new model to \emph{achieve term-level simplification and layperson-understandable sentence generation simultaneously}. 

\subsection{Our Contributions}
$\bullet$ \textbf{Dataset}. We manually annotate a new \underline{Med}ical \underline{Lan}guag\underline{e} simplification dataset named {\namedata}, which not only provides aligned sentence pairs but also offers term-level annotations.
The dataset consists of 12,801 training samples, 1,015 validation samples, and 1,016 testing samples.

\noindent$\bullet$ \textbf{Approach}. Following the human annotation procedure, we design a novel end-to-end \underline{D}ictionary-\underline{e}nhanced \underline{c}linical \underline{la}nguage simplifi\underline{er} (shorten for {\namemodel}), which consists of three parts, including \emph{a complex word locator, a neural lexical interpreter}, and \emph{a restricted syntactic polisher}. The locator is in charge of automatically recognizing medical jargon and abbreviations in the input sentences. The neural lexical interpreter aims to replace the located terms with appropriate professional expressions selected from a predefined dictionary. Finally, a syntactic polisher is implemented to simplify the modified sentences by the interpreter and further increases the readability and understandability of original sentences, which should be significantly helpful for users with low health literacy.

\noindent$\bullet$ \textbf{Baseline}. We compare the proposed {\namemodel} against \emph{nine baselines}, including a dictionary-based approach, a statistical machine translation approach, three neural machine translation approaches (i.e., Seq2Seq~\cite{bahdanau2015neural} and its two variants), a modified text summarization model PointerNet~\cite{NIPS2015_5866}, two state-of-the-art transformer-based pre-trained Seq2Seq models (i.e., T5~\cite{raffel2019exploring} and BART~\cite{lewis2019bart}), and PMBERT-MT that is built upon the pre-trained language model PubMedBERT~\cite{gu2020domain}, for validating the usability of the {\namedata} dataset and the effectiveness of our model. 
We also list {EditNTS}~\cite{dong2019editnts}, an approach in the general ATS domain, as a baseline.

\noindent$\bullet$ \textbf{Evaluation}. Different from bilingual translation tasks, our task is to translate professional medical language to layperson-understandable language. We not only require the translated results to be readable but also accurate and easily understandable. Thus, we design three new yet general evaluation metrics for the clinical simplification task. Besides, we still evaluate the results with commonly-used evaluation metrics, including BLEU~\cite{papineni2002bleu}, METEOR~\cite{lavie2007meteor}, ROUGE-L~\cite{lin2004rouge}, CIDEr~\cite{vedantam2015cider}, and SARI~\cite{xu2016optimizing}.

\section{Related Work}
\subsection{Medical Text Simplification Datasets}
We summarize the widely-adopted clinical-related text simplification datasets and make a comparison with our proposed {\namedata} from different angles as shown in Table~\ref{Tab:CompData}.

The first type of datasets focus on the normalization of abbreviations and acronyms by choosing proper explanations for them from a predefined dictionary, e.g., n2c2-track3~\cite{henry20202019} and CLEF~\cite{elhadad2013share}. The datasets are designed for a \emph{classification} problem and do not include any sentence-level polishing and complex phrase translation. Thus, the readability of the simplified sentences may still be low even after the term-level normalization.

The second type of datasets, including MSD~\cite{cao2020expertise}, CDSR~\cite{guo2020automated}, e-prescription~\cite{zheng2021work} and \cite{van2019evaluating}, focus on the style translation setting, which ignores the term-level simplification. 
The most similar dataset is the work~\cite{sakakini2020context}. Except for missing the term-level annotation, it is not publicly available and only contains a small number of annotated free-text sentence pairs, which cannot be used by deep learning models.
\subsection{Medical Text Simplification Method}
Medical text simplification is a sub-task of automatic text simplification (ATS)~\cite{laban2021keep,dong2019editnts,stahlberg2020seq2edits,paetzold2016unsupervised,paetzold2017lexical}, whose goal is to reduce the linguistic complexity of the original text to improve the readability. 
Besides increasing the readability, another target of the medical text simplification task is to accurately simplify professional medical terms. Existing approaches for medical text simplification can be roughly classified into two categories.

\emph{Dictionary-based approaches} rely on using the expert-curated medical dictionaries to simplify the professional medical sentences~\cite{kandula2010semantic, zeng2006exploring, zeng2007making, chen2017ranking,lalor2019improving} or link medical
terms with lay definitions~\cite{chen2018natural} and definitions in controlled vocabularies~\cite{polepalli2013improving}. These approaches are highly reliable yet cannot manage the case of polysemant, i.e., and a term can have multiple correct explanations under different cases. To solve this issue, \cite{sakakini2020context} utilizes a pre-trained language model to select the most possible answer and then replaces the selection with the located hard terms. This approach is an advanced version of the dictionary-based approach. However, this simple replacement may lead to a decrease in sentence readability, which further makes the simplified sentence still difficult to be understood by users or patients.


Researchers also try to borrow ideas from \emph{machine translation}, like aligning word embeddings between professional terms and daily expressions to achieve the translation~\cite{Kang2016A, Weng2018Mapping}, or further using back-translation procedures~\cite{weng2019unsupervised} and denoising autoencoder~\cite{pattisapu2020leveraging} to improve the simplification results under unsupervised conditions. 
For supervised approaches, PharmMT~\cite{li2020pharmmt} uses the Seq2seq~\cite{bahdanau2015neural} model pulsing a numerical checker to perform the simplification. \cite{devaraj2021paragraph} improves the BART\cite{lewis2019bart} model using the unlikelihood constraint~\cite{welleck2019neural} to penalize the model generating technical words. 
Although the readability of these methods is higher compared to the dictionary ones by directly modifying sentences, the accuracy of term-level translation cannot be guaranteed.
To address these problems, in the design of the {\namemodel}, we absorb the advantages of both machine translation and dictionary-based approaches to simplify clinical text from both sentence and term levels.

\section{{\namedata}: A New Benchmark Dataset}\label{sec:dataset}


\subsection{Data Collection}
The MIMIC-III database~\cite{johnson2016mimic} contains de-identified data from 58,976 ICU patient admissions, which includes several types of medical information such as demographics, medications, clinical notes, and so on.
We first select clinical notes from the NOTEEVENTS table of the MIMIC-III v1.4 dataset\footnote{\url{https://mimic.physionet.org/mimictables/noteevents/}} focusing on the following three sections: (1) History of present illness, (2) Brief summary of hospital course, and (3) Brief hospital course. These three sections contain thoughts and reasoning for the communication between clinicians, which are usually written with professional medical jargon and abbreviations. However, they still contain many plain sentences such as \emph{``She now also reports of being hunger.''}. To avoid translating them again, we design a heuristic feature-based sentence selection approach to filter out such sentences. In particular, we first tokenize each sentence into a set of words and then use a dictionary-based approach to match medical abbreviations. We also count the number of commonly-used English words within a given list\footnote{\url{https://www.ef.com/wwen/english-resources/english-vocabulary/top-3000-words/}}. 
Based on the length of the sentences (greater than 10), the ratio of medical abbreviations (smaller than 0.5), and the ratio of top-3000 words (greater than 0.1), we can automatically select candidate sentences. 
The sentence selection algorithm is summarized in Algorithm~\ref{alg:training_1}. 

\begin{algorithm}[!t]
\caption{\small Sentence Selection Algorithm} 
\label{alg:training_1}\small
\LinesNumbered
\KwIn{Target sentence $s$, top-3000 word set $T$, medical abbreviation set $A$}
\KwOut{Selected sentence set}
Tokenize $s$ into words $[w_1,...,w_n]$\;
\For{$i=1$ to $n$}{
    \If{$w_i\in A$}{
        $abb$ = $abb$ + 1\;
    }
    {$w_i^{\prime} = lemmatize(w_i)$\;}
    \If{$w_i^{\prime} \notin T$}{
        $unc = unc + 1$\;
    }
}
\eIf{$n<10$ or $\frac{unc+abb}{n}>0.5$ or $\frac{unc+abb}{n}<0.1$}
{
    \textbf{return} False\;
    }
{ \textbf{return} True;}

\end{algorithm}

\subsection{Data Annotation}
After we collect a set of source sentences, the next step is to annotate them. 
However, annotating medical sentences is different from creating a parallel bilingual translation corpus. The medical language translation task aims to ``translate'' professional and clinical jargon to layperson-understandable language, which is still from the same language but uses different expressions. Besides, annotating bilingual translation data focuses on readability and accuracy. Except for those two perspectives, annotating medical sentences also considers understandability, which is from the perspective of patients or customers. Based on the above guidance, 
we invited six researchers to annotate the data. All of them are familiar with medical data. For each sentence, there are two extra senior researchers holding the Doctor of Medicine (M.D.) degree to check the annotation quality. For the translated sentences with low quality, senior researchers need to re-translate them.

\begin{figure}[t]
\centering
\includegraphics[width=0.45\textwidth]{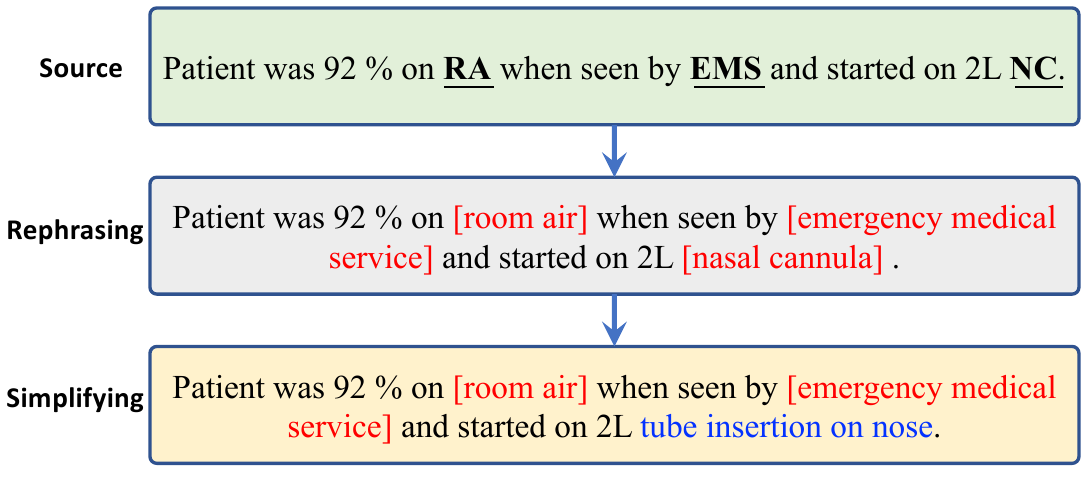}
\vspace{-0.1in}
\caption{An example of annotating a sentence by a worker using two steps, i.e., rephrasing and simplifying. In the rephrasing step, three abbreviations are replaced by full forms. In the simplifying step, the full form ``nasal cannula'' is replaced by ``tube insertion on nose''.}
\label{image:annotation}
\vspace{-0.2in}
\end{figure}

The purpose of this work is to create a benchmark for the automated clinical language understanding task, which is not only used for training translation models but also for fair evaluation of different approaches. Thus, we set different requirements for workers when annotating the training data and validation/testing data. In general, they use two steps when annotating each source sentence. The first step is to paraphrase the abbreviations with the whole words or phrases. For each abbreviation, there may be several full forms. For example, ``\emph{TLC}'' has two full forms\footnote{\url{https://medical-dictionary.thefreedictionary.com/TLC}}. One is ``\emph{thin-layer chromatography}'', and the other is ``\emph{total lung capacity}''. Therefore, it is important for workers to understand the context in which the abbreviation or term has been used. Note that we do not provide a dictionary for workers, and they search the full expressions on the Internet if they are not sure about the abbreviations. The second step is to use simple words to replace professorial medical expressions. Take the word ``\emph{hematocrit}''\footnote{\url{https://www.mayoclinic.org/tests-procedures/hematocrit/about/pac-20384728}} as an example, which means the ratio of the volume of red blood cells to the volume of the whole blood. If we use the expression, \emph{``the proportion of red blood cells in the blood''}, it is much more understandable for patients compared with directly using professional clinical jargon. An example in Figure~\ref{image:annotation} illustrates the annotating procedure.

When annotating the training data, workers are asked to return the final understandable sentences, i.e., the simplified ones, which will be checked by experts again to guarantee the annotation quality. For validation and testing data, we require workers to return both rephrasing and simplifying forms for each source sentence. 

Note that for all the training, validation, and testing data, there is a special case that we do not need to translate the source sentence. For example, it is easy to understand the sentence ``\emph{He had a set of surveillance blood cultures drawn last week, which were negative.}''. These sentences are extremely useful when training an understandable translation model because they can be considered as important indicators for guiding model learning.  
In the validation and testing data, another special case is that the sentence may not be simplified any more. For example, the source sentence is ``\emph{She also had subjective {SOB} with {CXR} suggesting fluid overload.}'', and the rephrased and simplified sentences are the same, which is ``\emph{She also had subjective [{shortness of breath}] with [{chest x-ray}] suggesting fluid overload.}''.

\begin{table}[!htb]
\centering
\resizebox{0.48\textwidth}{!}{
\footnotesize
\begin{tabular}{|l|c|}
\hline
\# of tokens in the source sentences & 14,780\\\hline
\# of tokens in the target sentences & 14,278\\\hline
\# of overlapped tokens between source \& target  & 12,501\\\hline
Avg. length of the source sentences  & 20.6\\\hline
Avg. length of the target sentences  & 24.0\\\hline
Avg. \# of abbreviations in validation \& testing sets  & 1.2\\\hline
\end{tabular}
}
\vspace{-0.1in}
\caption{{\namedata} data statistics.}
\label{Table:data}
\vspace{-0.2in}
\end{table}

\subsection{Dataset Statistics}
The {\namedata} dataset contains 12,801 sentences for training, 1,015 sentences for validation, and 1,016 sentences for testing. Table~\ref{Table:data} shows the statistics of the {\namedata} dataset, which are different from those of traditional machine translation datasets.
First, the way of annotation is different, as we discussed in the previous section. Second, there are a large number of overlapped tokens between source and target sentences, which is also different from traditional machine translation. These differences make our task unique and challenging.

\section{{\namemodel}: An Effective Approach}
\begin{figure*}[h]
\centering
\includegraphics[width=0.75\textwidth]{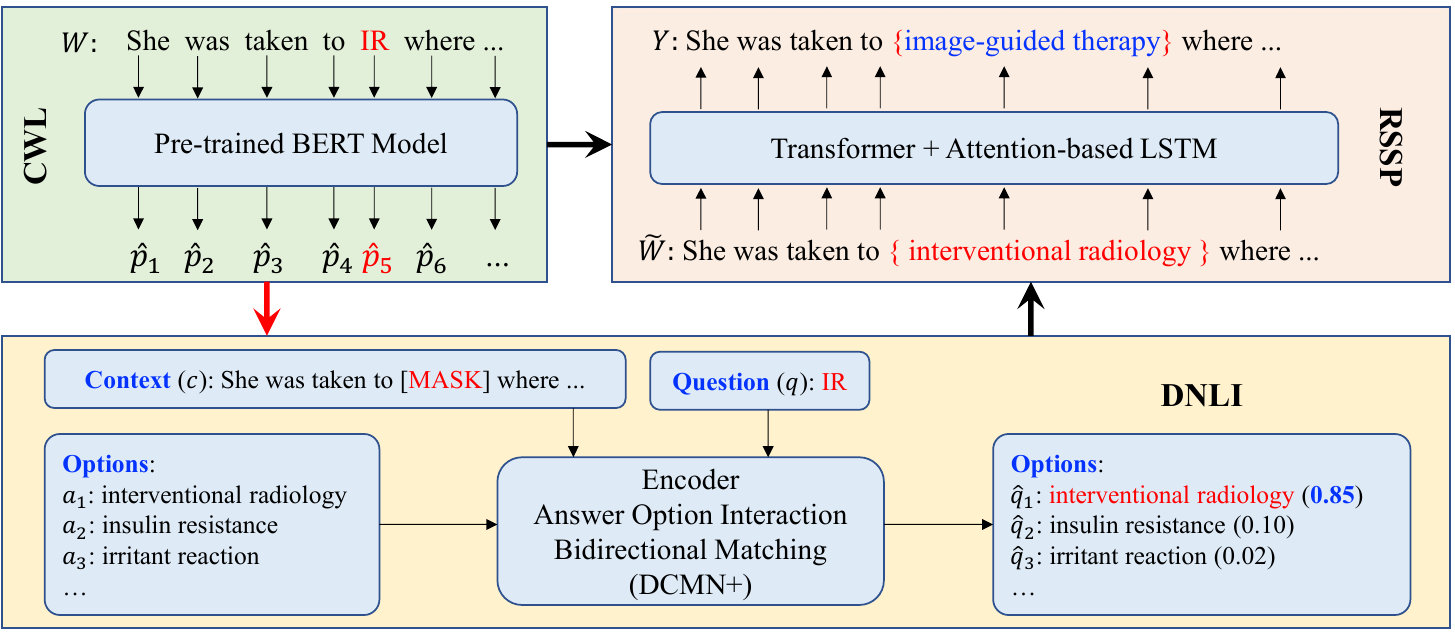}
\vspace{-0.1in}
\caption{Overview of the proposed {\namemodel} model.}
\label{image:model}
\vspace{-0.2in}
\end{figure*}

Motivated by the annotation steps, we propose an effective end-to-end model named {\namemodel} for the automated clinical language simplification task. In particular, we collect a medical dictionary to estimate the possible full-term expressions of abbreviations in the input sentences. The proposed model is shown in Figure~\ref{image:model}, which consists of three main modules, i.e., a complex word locator (CWL), a dictionary-based neural lexical interpreter (DNLI), and a restricted syntactic simplification polisher (RSSP). 

Given a tokenized professional medical sentence $W = [w_1,w_2,\cdots, w_n]$, where $n$ denotes the number of tokens, the locator aims to dig out possible phrases that need to be simplified or translated. In the neural lexical interpreter, the chosen phrases will be replaced with full-term expressions selected from the medical dictionary. Finally, the replaced sentence will pass the polisher to generate the final output $Y$. These three parts tightly work together and enhance each other. Next, we introduce the design details of each module, respectively.

\subsection{Complex Word Locator (CWL)}
As we discussed in Section~\ref{sec:dataset}, clinical language understanding is different from the traditional machine translation task, and we only need to modify a part of professional medical jargon in the input sentence $W = [w_1,w_2,\cdots, w_n]$ and keep the remaining words. Towards this end, we design a complex word locator to find out the tokens that need to be modified. Note that in the annotated dataset, we have such indicators that which tokens are modified.
In particular, we use a pre-trained BERT model~\cite{devlin2018bert} with PubMed data, i.e., PubMedBERT~\cite{gu2020domain}, to learn a representation for each input token $w_i$, and a fully-connected layer (FC) followed by the softmax function is used to calculate the probability of each token to be modified or not. Let $\mathbf{\hat{p}}_i$ denote the binary probability vector, and we have
\begin{align}
    \mathbf{\hat{p}}_i = \text{softmax}(\text{FC}(\text{PubMedBERT}(w_i))).
\end{align}

Let $\mathbf{p}_i \in \{0, 1\}^2$ be the ground truth vector for the $i$-th token and $\mathcal{L}_p$ denote the average cross-entropy (CE) loss function, i.e., 
\begin{align}
    \mathcal{L}_p = \frac{1}{n} \sum_{i=1}^n \text{CE}(\mathbf{p}_i, \mathbf{\hat{p}}_i).
\end{align}

\subsection{Dictionary-based Neural Lexical Interpreter (DNLI)} 
When workers annotate the source sentence, the first step is to rephrase the abbreviations. Using the designed locator, {\namemodel} is able to identify the possible abbreviations. Then we use a dictionary-based neural lexical interpreter to automatically substitute it with an appropriate full form for each located token. Note that if the located token does not have any full-term expression, we will keep it in the sentence. Some abbreviations have several full-term expressions, and the proposed neural lexical interpreter will automatically choose one with the highest probability based on the context information, i.e., the remaining tokens in the sentence. The details of constructing the dictionary can be found in Section~\ref{sec:dictionary}.

To recover the most appropriate full version of located token $w_i$, we borrow the idea of DCMN+~\cite{zhang2020dcmn+} to design the dictionary-based neural lexical interpreter. Specifically, we mask the located token $w_i$ from the input sentence, i.e., the masked sentence is $W^\prime = [w_1, \cdots, w_{i-1}, [\text{MASK}], w_{i+1}, \cdots, w_n]$, which can be considered as the context information. Then the located token $w_i$ can be regarded as the question, and all the full terms $A= \{a_1, \cdots, a_m\}$ are considered as options, where $m$ is the number of possible full versions. The goal is to select the best candidate $\hat{a}_j$ from the options $A$ when given the context vector $W^\prime$ and the question $w_i$. 

We first encode $W^\prime$ with PubMedBERT and denote the encoded vector as $\mathbf{c}$. Similarly, the question $w_i$ can also be mapped to a vector $\mathbf{q}$, and each option $a_j$ can be converted to a vector $\mathbf{a}_j$. Then using answer option interaction and bidirectional matching modules in DCMN+, we can estimate the probability of each full-term expression to be selected, which is denoted as $\mathbf{q} = [q_1, \cdots, q_m]$. 

Note that there may be multiple located tokens in a sentence, and we will generate the corresponding number of $\{$\emph{context, question, options}$\}$ pairs. Besides, when annotating the {\namedata} dataset, the annotators use square brackets ``[]'' to indicate the correct full term expressions as shown in Figure~\ref{image:annotation}. Thus, there are ground truths that are denoted as $\mathbf{\hat{q}}$ for $\{$\emph{context, question, options}$\}$ pairs.
Assume that there are $l$ located tokens in the input sentence, then we will have the following loss function:
\begin{equation}
    \mathcal{L}_q = \frac{1}{l} \sum_{j=1}^l \text{CE}(\mathbf{q}_j, \mathbf{\hat{q}}_j).
\end{equation}
The output of the dictionary-based neural lexical interpreter is a new sentence by replacing located tokens in the locator with the best candidates selected from the dictionary, which is denoted as $\tilde{W} = [\tilde{w}_1, \cdots, \tilde{w}_{n^\prime}]$, where $n^\prime$ is the number of tokens in the new sentence $\tilde{W}$.

\subsection{Restricted Syntactic Simplification Polisher (RSSP)}
The second step of annotation is to polish the sentences to make them more fluent, simple, and understandable by patients with low health literacy. An easy way is to directly ``translate'' the new sentence $\tilde{W}$ to the target sentence with a neural machine translation model. However, as we mentioned before, the clinical language understanding task is different from traditional neural machine translation, which aims to make the professional clinical jargon understandable by patients. In fact, most of the tokens in the sentences can be kept and do not translate again, and we only polish the terms labeled by the locator and replace them with the interpreter. Thus, this is a \emph{partial translation task}.

First, we add special markers to $\tilde{W}$ to indicate the tokens to be polished, i.e., $\tilde{W} = [\tilde{w}_1, \cdots, \{\tilde{w}_i, \cdots, \tilde{w}_j\},\cdots, \tilde{w}_{n^\prime}]$, where the tokens from $\tilde{w}_i$ and $\tilde{w}_j$ will be ``translated'' by the polisher. The polisher first encodes the input sentence $\tilde{W}$ via Transformer~\cite{vaswani2017attention}, i.e., $
    \mathbf{v}_i = \text{transformer}(w_i |w_1, \cdots, w_{n^\prime})
$, where $\mathbf{v}_i$ is the representation of the $i$-the token. Then an attention-based LSTM~\cite{hochreiter1997long,sutskever2014sequence,bahdanau2015neural,DBLP:journals/corr/abs-2110-14844} is used as the decoder to generate the $t$-th word $y_t$ within braces. Let $\mathbf{h}_t$ represent the hidden state outputted by the decoder LSTM, and then we can obtain the weighted context vector $\mathbf{s}_t$ using the attention mechanism as follows:
\begin{align*}
        \mathbf{s}_t = \sum_{i=1}^n \alpha_{ti} \mathbf{v}_i, \alpha_{ti}  = \frac{\exp({o}_{ti})}{\sum_{j=1}^n(\exp({o}_{tj}))}, {o}_{ti}  = \mathbf{h}_t \mathbf{v}_i^\top.
\end{align*}
Finally, we can obtain the probability of the $t$-th word:
$
    \mathbf{r}_t
    = \text{softmax}(\mathbf{W}[\mathbf{h}_t; \mathbf{s}_t] + \mathbf{b}),
$
where $\mathbf{W}$ and $\mathbf{b}$ are two parameters. 
Assume that the $k$-the element of $\mathbf{r}_t$ corresponds to the truth token, and then we have the loss of the polisher as follows:
\begin{equation}
    \mathcal{L}_r = -\frac{1}{j-i+1}\sum_{t=i}^j \log(\mathbf{r}_t[k]).
\end{equation}

\subsection{Training}
The proposed {\namemodel} is an end-to-end model, and we can train the model using the following loss function:
\begin{equation}
    \mathcal{L} = \mathcal{L}_p + \mathcal{L}_q + \mathcal{L}_r.
\end{equation}
In the evaluation stage, we can run these three modules one by one to generate the final understandable sentences.

\section{Experiments}

\subsection{Experimental Setups}\label{sec:dictionary}
\textbf{Dictionary Construction}. We construct the mapping dictionary by collecting medical abbreviations and their corresponding full forms from the book~\cite{DORLAND2016dictionary} and online sources, including Charleston Area Medical Center\footnote{\url{https://bit.ly/3uMexm6}}, Taber's Medical Dictionary\footnote{ \url{https://bit.ly/3uR4DzL}}, your dictionary\footnote{\url{https://bit.ly/3fkuNEu}}, MedicineNet\footnote{\url{https://bit.ly/3oqUw27}}, and Wikipedia\footnote{ \url{https://bit.ly/3yg7l3K}}.
\begin{table*}[!htb]
\centering
\resizebox{\textwidth}{!}{
\begin{tabular}{|c|c|c|c|c|c|c|c|c|c|c|c|c|}
\hline
\textbf{Model}            & \textbf{BLEU-1}                         & \textbf{BLEU-2}  & \textbf{BLEU-3}  & \textbf{BLEU-4}  & \textbf{BLEU} & \textbf{METEOR} & \textbf{ROUGE-L} & \textbf{CIDEr} & \textbf{SARI} & \textbf{HIT}    & \textbf{CWR}    & \textbf{AScore} \\ \hline
Dictionary & 0.7158                       & 0.6364 & 0.5684 & 0.5076 & 0.6070 & 0.3933 & 0.7308 & 4.2037 & 37.3391 & 0.5572 & 0.6407 & 0.5948 \\ \hline
Moses      & 0.7880                       & 0.7130 & 0.6530 & 0.6016 & 0.6889 & 0.4237 & 0.8188 & 5.1046 & 51.6827 & 0.6823 & 0.7543 & 0.6859 \\ \hline
Seq2seq    & 0.7136                       & 0.6322 & 0.5969 & 0.5160 & 0.6147 & 0.3533 & 0.7609 & 4.1299 & 46.1328 & 0.7388 & 0.7980 & 0.6648 \\ \hline
Seq2seq-   & 0.5066                       & 0.3315 & 0.2373 & 0.1787 & 0.3135 & 0.1859 & 0.4948 & 1.2670 & 24.5346 & 0.6427 & \textbf{0.8367} & 0.4070 \\ \hline
Seq2seq-S  & 0.7180                       & 0.6386 & 0.5778 & 0.5267 & 0.6153 & 0.3604 & 0.7683 & 4.2635 & 46.5085 & 0.7331 & 0.7953 & 0.6630 \\ \hline
PointerNet & 0.6870                       & 0.5904 & 0.5158 & 0.4541 & 0.5618 & 0.3338 & 0.7285 & 3.9458 & 42.2857 & 0.6414 & 0.7555 & 0.5949 \\ \hline
EditNTS    & 0.8213                       & 0.7801 & 0.7452 & 0.7132 & 0.7649 & 0.4674 & 0.7401 & 5.9508 & 62.6036 & 0.6405 & 0.6915 & 0.7116\\ \hline
BART       & 0.7148                       & 0.6755 & 0.6396 & 0.6060 & 0.6590 & \textbf{0.5320} & 0.7616 & 4.9783 & 70.3058 & 0.5266 & 0.7311 & 0.6191 \\ \hline
T5         & 0.7223                       & 0.6812 & 0.6445 & 0.6103 & 0.6646 & 0.5305 & 0.7645 & 5.0629 & 71.3255 & 0.5262 & 0.7342 & 0.6220 \\ \hline
BERT-MT    & 0.8003                       & 0.7428 & 0.6952 & 0.6531 & 0.7228 & 0.4566 & 0.8218 & 5.3293 & \textbf{72.2260} & 0.7808 & 0.7358 & 0.7417 \\ \hline
 \rowcolor{lavendermist} {\namemodel} & \textbf{0.8624}               & \textbf{0.8291} & \textbf{0.8004} & \textbf{0.7737} & \textbf{0.8165} & 0.5290 & \textbf{0.8894} & \textbf{6.7212} & 70.8583 & \textbf{0.7986} & 0.7328 & \textbf{0.7983} \\ \hline
 \rowcolor{lavendermist}$\uparrow$    & +5.0\% & +6.3\% & +7.4\% & +8.5\% & +6.7\% & -0.5\% & +8.2\% & +26.1\% & -1.9\% & +2.2\% & -12.4\% & +7.6\% \\\hline
\end{tabular}
}
\caption{Performance evaluation of all the baselines with different metrics. $\uparrow$ denotes the percentage of performance gain compared with the best baselines.}
\label{Table:exp}
 \vspace{-0.1in}
\end{table*}

\smallskip
\noindent\textbf{Baselines}.
We use the following approaches as baselines:  
a simple term replacement approach named \emph{Dictionary-based model}, a statistical machine translation (SMT) system
\emph{Moses}\footnote{\url{http://www.statmt.org/moses/}}, 
neural machine translation approaches, including \emph{Seq2Seq}~\cite{bahdanau2015neural} and its two variants \emph{Seq2Seq}$-$ and \emph{Seq2Seq-S}, a modified version of the pointer network~\cite{NIPS2015_5866} \emph{PointerNet}, state-of-the-art language models \emph{BART}~\cite{lewis2019bart} and \emph{T5}~\cite{raffel2019exploring}, and a modified BERT-based simplifier \emph{BERT-MT}.
We also include \emph{EditNTS}~\cite{dong2019editnts} -- a method proposed for the general ATS -- as a baseline. Although there are other more advanced methods~\cite{martin2020muss, martin2019controllable, maddela2020controllable} in the general ATS domain, they all require special grammar level information like the part of speech and syntactic tree information, which is hard to obtain for our clinical domain dataset. Thus, they are not compared in our experiments.

\noindent\textbf{Parameter Settings}. 
For the dictionary method, we use the pre-constructed dictionary as the same as the {\namemodel} model.
For the statistical model Moses, we follow the training procedure listed on the User Manual and Code Guide file\footnote{\url{http://www.statmt.org/moses/manual/manual.pdf}}. 
For neural machine translation models and text summarization baseline, we all conduct a grid search to find the optimal parameters. 
For Seq2Seq, Seq2Seq$-$, Seq2Seq-S, and PointerNet, the hidden size is set to 256 for both encoder and decoder by greedy search, and the learning rate is set to $1e-3$. 
We use Adam~\cite{kingma2014adam} as the optimizer. Tokenization is performed using NLTK word tokenizer~\cite{bird2009natural}. The early stop is also applied by checking the BLEU score~\cite{papineni2002bleu} on the validation set, and the training batch size is set to 30.

For EditNTS, we use the original default parameter setting with the learning rate of $1e-3$ for the Adam optimizer. 
For the BERT-MT model, the hidden size is the same as that of PubMedBERT, which is 786. We also use the default AdamW optimizer used by PubMedBERT with the learning rate as $5e-5$, the warm-up method, the default PubMedBERT vocabulary, and tokenization are applied. For BART and T5, the setting of the optimizer and training procedure is the same as the BERT-MT.

For the proposed {\namemodel}, the locator is based on PubMedBERT to perform token-level classification, and we use the default setting of PubMedBERT to train the locator.
For the dictionary-based neural interpreter, we use the same parameter setting as~\cite{zhang2020dcmn+}. The max size of the answers is set to 8. The maximum length of the input sentence is set to 64 during training. The learning rate is set to $5e-5$ with ten epochs, and an early stop is adopted.
For the restricted polisher, the setting is the same as the BERT-MT model except for the restricted translation setting. PubMedBERT is applied with an LSTM decoder that has the same hidden size.

In the evaluation stage, the same NLTK word tokenizer is applied as baselines to break the sentences into words for calculating the scores for a fair comparison. All models are trained on Ubuntu 16.04 with 128GB memory and an Nvidia Tesla P100 GPU.

\smallskip
\noindent
\textbf{Evaluation Metrics}.
We use BLEU~\cite{papineni2002bleu}, METEOR~\cite{lavie2007meteor}, ROUGE-L~\cite{lin2004rouge}, and CIDEr~\cite{vedantam2015cider} scores as the evaluation metrics, which are widely-used for the machine translation task. 
Besides, we use SARI~\cite{xu2016optimizing}, which is specially designed for the general ATS task. It combines the $n$-gram evaluation method of the BLEU score and rewards the replacement of the input words. However, it fails to cover the accuracy requirement of the professional medical term simplification. 

Since our task is different from traditional machine translation and ATS tasks, directly applying existing evaluation metrics cannot fairly evaluate the performance of different models.
Since our task is different from traditional machine translation tasks, directly applying existing evaluation metrics cannot fairly evaluate the performance of different models. 
Here, we use an example in Figure~\ref{fig:example} to demonstrate the failure of existing evaluation metrics, such as the BLEU score. If we directly copy the original source sentence as the answer, a very high BLEU score can be obtained, which is 0.85. However, the critical term ``\emph{pt}'' is not translated. Without translating this term, patients or customers may not totally understand the meaning of this sentence.
Thus, it is necessary to design task-specific metrics.
\begin{figure}[!htb]
 \centering
\begin{minipage}[h]{0.43\textwidth}
\fbox{
\small
\parbox{\textwidth}{
{\tt{
\textbf{Source}: vascular saw the \textcolor{blue}{pt} and did not feel that there was an acute need for an invasive procedure.\\
\textbf{Target}: vascular saw the \textcolor{red}{patient} and did not feel that there was an acute need for an invasive procedure.
}}
}
}
\end{minipage}
\caption{Example of the failure of existing metrics.}\label{fig:example}
\end{figure}

$\bullet$ \emph{Hit Ratio (HIT)}. A key challenge of medical language translation is to translate professional medical jargon into layperson-understandable words. Let $n_p$ denote the number of professional medical terms in a source sentence and $n_t$ be the number of correctly translated terms in the target. We then have the HIT ratio, which is $HIT = \frac{n_t}{n_p}$.
    
$\bullet$ \emph{Common Word Ratio (CWR)}. To evaluate the simplicity of the translated sentences, we follow the work of Dale–Chall readability~\cite{mcclure1987readability} to calculate the common word ratio for each output sentence. 
We first lemmatize each word of the output. If the lemmatized form is in the top-3000 commonly-used English words, then it is a common word. Otherwise, it is not a common word. Let $n_c$ denote the number of common words in the translated sentence, and $n$ represents the length of the translated sentence. The common word ratio score is $CWR=\frac{n_c}{n}$.

$\bullet$ \emph{Aggregation Score (AScore)}. The quality of the translated sentences is not only decided by the readability (BLEU) but also related to the correctness (HIT) and simplicity (CWR). Among these three perspectives, readability and correctness should be more important than simplicity. Thus, we design a new score to model them jointly, which is
    \begin{align*}
        Ascore = \frac{1+\alpha^2+\beta^2}{\frac{\alpha^2}{BLEU} + \frac{\beta^2}{HIT} + \frac{1}{CWR}},
    \end{align*}
    where $\alpha$ and $\beta$ are parameters for controlling the importance of BLEU and HIT scores. If any of the three metrics is 0, then it will be added to a very small number such as $10^{-8}$ to avoid AScore being 0. We take $\alpha=2$ and $\beta=1.5$ in our experiments.


\subsection{Experimental Result Analysis}
Table~\ref{Table:exp} shows the experimental results of all baselines and {\namemodel} on different evaluation metrics. 
The Dictionary method is the simplest approach, but its performance is not the worst compared with other baselines in terms of transitional machine translation evaluation metrics such as BLEUs. The reason is that there are many overlapping tokens in both source and target sentences, which is the main difference between the traditional machine translation task and our clinical language understanding task. These results also confirm that we need to design new evaluation metrics for this new task.

General neural network-based approaches, including Seq2Seq, its variants, and PointerNet, have a relatively low BLEU score. The reason is that the labeled data is insufficient for them to train a powerful translation model from scratch. 
For the general ATS method EditNTS, we can observe a relatively high score for BLEU. EditNTS is good at keeping the information. However, due to the lack of external knowledge support, the HIT score is still not satisfactory enough, proving the importance of the DNLI module. In addition, EditNTS is designed for the general ATS domain. Many attributes are not suitable for our task, which can also contribute to the bad performance.
For the transformer-based pre-trained sequence to sequence models T5 and BART, the BLEU score is relatively high but with a much lower HIT score compared to Seq2Seq and PointerNet. The unsatisfied results may be related to the domain shifting problem since T5 or BART models are not pre-trained for the medical text-domain. 
On the contrary, BERT-MT conducting pre-training on a large medical language corpus significantly increases the ability of model learning. Hence, the performance of BERT-MT is the best among all the baselines. However, the proposed {\namemodel} mimics the human annotation procedure and employs a mapping dictionary with a novel model design, which leads to achieving the best performance compared with all the baselines. 

Using the pre-trained language models is helpful to attain a higher SARI score because the pre-training technique can increase the models' generalization ability and benefit the word replacement rewards of the SARI metric. However, SARI does not focus on the accuracy of simplifying professional medical abbreviations, which makes it unable to comprehensively and fairly evaluate the results.


From the view of the HIT score, we can find a sufficient gap between generation-based methods (including Seq2Seq, Seq2Seq-S, BERT-MT, and {\namemodel}) and other methods. To achieve a high HIT score, the accurate recognition of abbreviations is necessary. Moreover, the models should understand the context correctly, which is an advantage of neural network-based models. Besides, another requirement is the generation ability. The use of a reference mechanism probably limits the generation ability of the PointerNet model, and thus, it does not achieve a high HIT score.

The CWR score can reflect the simplicity of sentences in a scene. However, we should notice that the higher CWR scores do not mean better performance. The reason is that translating professional medical terms will inevitably result in some uncommon words. Thus, we should attach less importance to the CWR score when calculating the comprehensive rank.

The top 3 models in the view of AScore are {\namemodel}, BERT-MT, and EditNTS, which is reasonable. The AScore attaches the highest importance to BLEU, followed by the HIT and CWR scores. Using the harmonic mean can make sure that we penalize the tendency of going overboard on one subject and guarantee good general performance.  

\section{Conclusion}
This paper aims to benchmark a new, challenging, yet practical task of automated clinical language simplification by constructing a high-quality {\namedata} dataset and proposing a new model {\namemodel} that mimics the human annotation procedure. We conducted experiments on the annotated {\namedata} dataset by implementing nine strong baselines against {\namemodel}. Experimental results confirmed the utility of the constructed {\namedata} dataset and the effectiveness of the proposed {\namemodel} for addressing the automated clinical language simplification task.

\section{Broader Impacts}
Current health care service and health information technology (HIT) system design strive to provide accessible ways for patients to be engaged in their own care and make informed decisions. One instance is to make personal health records accessible to patients through patient portals (the electronic personal health record systems connected to organizations’ electronic health record systems). However, only providing access to health records is insufficient to fully empower patients. Patients may struggle to understand those records due to such reasons as low health literacy, unfamiliarity with medical jargon and clinical abbreviations, or difficulty in understanding the complex structure of medical language. Currently, only a few studies target this practical issue. In this paper, we not only provide a human-annotated dataset, design a new model but also propose three evaluation metrics for benchmarking the clinical language simplification task. 
Our work will expedite research in multiple domains, including but not limited to, natural language processing, machine learning, and healthcare informatics.

\section{Ethical Consideration}
The original data of our study are directly extracted from the MIMIC-III database~\cite{johnson2016mimic}, where all private health information was de-identified. The MIMIC-III database was performed under Health Insurance Portability and Accountability Act (HIPAA) standards, which require the removal of all the identifying data elements in the list of HIPAA (e.g., name, phone number, address, and so on). Thus, this is no privacy issue for the data that we use. 
When annotating the dataset, all annotators submitted all required consent forms. 
Since this work only focuses on simplifying clinical text, and no additional identified and private information is added. As a result, the protection of privacy is preserved. For disseminating our dataset to be publicly available, we will follow the same requirement of the MIMIC-III data. In other words, the requester must complete a recognized training for protecting human research participants and sign an agreement to protect the data privacy following the requirement of the PhysioNet\footnote{\url{https://physionet.org/}}~\cite{goldberger2000physiobank}. 




\bibliography{refer}
\bibliographystyle{acl_natbib}
\appendix
\section*{Appendix}
\section{Sentence Selection}\label{app:sen_sel}
The sentence selection algorithm is summarized in Algorithm~\ref{alg:training}. For each tokenized word $w_i$, we first check whether it belongs to the medical abbreviation set $A$, and $abb$ in line 4 denotes the current number of medical abbreviations in the target sentence $s$. If $w_i$ is not a medical abbreviation, then we will check whether the lemmatized $w_i$ belongs to the top-3000 word set $T$. $unc$ in line 8 represents the current number of words that are neither medical abbreviations nor commonly-used words. Finally, based on the predefined criteria, the algorithm can automatically decide to keep or remove the target sentence. 
\begin{algorithm}[h]
\caption{Sentence Selection Algorithm} 
\label{alg:training}
\LinesNumbered
\KwIn{Target sentence $s$, top-3000 word set $T$, medical abbreviation set $A$}
\KwOut{Selected sentence set}
Tokenize $s$ into words $[w_1,...,w_n]$\;
\For{$i=1$ to $n$}{
    \If{$w_i\in A$}{
        $abb$ = $abb$ + 1\;
    }
    {$w_i^{\prime} = lemmatize(w_i)$\;}
    \If{$w_i^{\prime} \notin T$}{
        $unc = unc + 1$\;
    }
}
\eIf{$n<10$ or $\frac{unc+abb}{n}>0.5$ or $\frac{unc+abb}{n}<0.1$}
{
    \textbf{return} False\;
    }
{ \textbf{return} True;}
\end{algorithm}

\section{Restricted Translating}\label{app:restricted}
The restricted translating algorithm is summarized in Algorithm~\ref{alg:tran}. 
In line 5, we first check whether the algorithm meets the special token to decide entering the copy or translating state.
Lines 6-8 illustrate the process of copy state, i.e., copying the original input and updating the state of the translator.
Lines 10-15 illustrate the translating state, and the translator model will use its own output to update the state and copy it to the answer $Y$. 
\begin{algorithm}
\caption{Partial Translation}
\label{alg:tran}
\LinesNumbered
\KwIn{Input Sentence $\tilde{W}$, Translation Model $Tran$}
\KwOut{Input Sentence $Y$}
$W$ = Tokenizer($\tilde{W}$)\;
$i = 0$\;
$Y$ = []\;
\While{Not complete generating $Y$}
{
    \eIf{In Copy Stage}
    {
        $Y$.append($W[i]$)\;
        $Tran$.next\_state($W[i]$)\;
        $i++$\;
    }
    {
        \While{Not Finish Translating}
        {
            $y$ = $Tran$.next\_word()\;
            $Y$.append($y$)\;
            $Tran$.next\_state($y$)\;
        }
        Skip $i$ to the next copy word\;
    }
    
}
return $Y$\;
\end{algorithm}



\section{Baselines \& Parameter Settings}\label{app:para_set}
\noindent\textbf{Baselines}. 
\emph{Dictionary-based model} means randomly select one full-term expression for each located token by the locator and directly replace the terms in the sentence as the output. 
\emph{Moses} is a widely-used statistical machine translation (SMT) system. 

Neural machine translation approaches include \emph{Seq2Seq}~\cite{bahdanau2015neural} and its two variants, i.e., \emph{Seq2Seq}$-$ without using the attention mechanism in the decoder and \emph{Seq2Seq-S} that shares the embedding space of encoder and decoder models.
\emph{PointerNet} is a modified version of the pointer network~\cite{NIPS2015_5866} by adding a generating/referring option to the model. In the referring mode, the model acts as a general pointer network. However, in the generating mode, the model acts like a normal Seq2Seq model. 
For general ATS methods, we select the EditNTS~\cite{dong2019editnts}, which is based on the sentence modification operations. 
For transformer-based models, we include
\emph{BART}~\cite{lewis2019bart}, which is a pre-trained transformer autoencoder framework and designed for natural language generation tasks; and \emph{T5}~\cite{raffel2019exploring} that is also a transformer autoencoder framework proposed by Google. Compared to \emph{BART}, \emph{T5} contains more advanced pre-trained tasks and has been proved to be a powerful framework on many natural language generations and understanding tasks.
The last baseline is \emph{BERT-MT}, which contains a BERT encoder and a LSTM decoder. It is similar to the polisher but directly translates the original inputs to the targets. 

\smallskip
\noindent\textbf{Parameter Settings}. For the statistical model Moses, we follow the training procedure listed on the User Manual and Code Guide file\footnote{\url{http://www.statmt.org/moses/manual/manual.pdf}}. 
For the dictionary method, we use the pre-constructed dictionary as the same as the {\namemodel} model.
For neural machine translation models and text summarization baseline, we all conduct a grid search to find the optimal parameters. 
For the EditNTS, we use the default original setting with the learning rate of $1e-3$ with Adam optimizer. The dimension setting is as same as the original work, a 200 dimension bi-direction RNN.
For the BERT-MT model, the hidden size is the same as that of PubMedBERT, which is 786. We also use the default AdamW optimizer used by PubMedBERT with the learning rate as $5e-5$, the warm-up method, the default PubMedBERT vocabulary, and tokenization are applied. For BART and T5, the setting of the optimizer and training procedure is the same as the BERT-MT.

Finally, for Seq2Seq, Seq2Seq$-$, Seq2Seq-S, and PointerNet, the hidden size is set to 256 for both encoder and decoder by greedy search, and the learning rate is set to $1e-3$. 
We use Adam~\cite{kingma2014adam} as the optimizer. Tokenization is performed using NLTK word tokenizer~\cite{bird2009natural}. The early stop is also applied by checking the BLEU score~\cite{papineni2002bleu} on the validation set, and the training batch size is set to 30.

For the proposed {\namemodel}, the locator is based on PubMedBERT to perform token level classification, and we use the default setting of PubMedBERT to train the locator.
For the dictionary-based neural interpreter, we use the same parameter setting as~\cite{zhang2020dcmn+}. The max size of the answers is set to 8. The maximum length of the input sentence is set to 64 during training. The learning rate is set to $5e-5$ with 10 epochs, and an early stop is adopted.
For the restricted polisher, the setting is the same as the BERT-MT model except the restricted translation setting. PubMedBERT is applied with an LSTM decoder that has the same hidden size.

In the evaluation stage, the same NLTK word tokenizer is applied as baselines to break the sentences into words for calculating the scores for a fair comparison. All models are trained on Ubuntu 16.04 with 128GB memory and an Nvidia Tesla P100 GPU.
\section{Experimental Results}\label{app:exp}

\begin{figure}[!htb]
\centering
\includegraphics[width=0.4\textwidth]{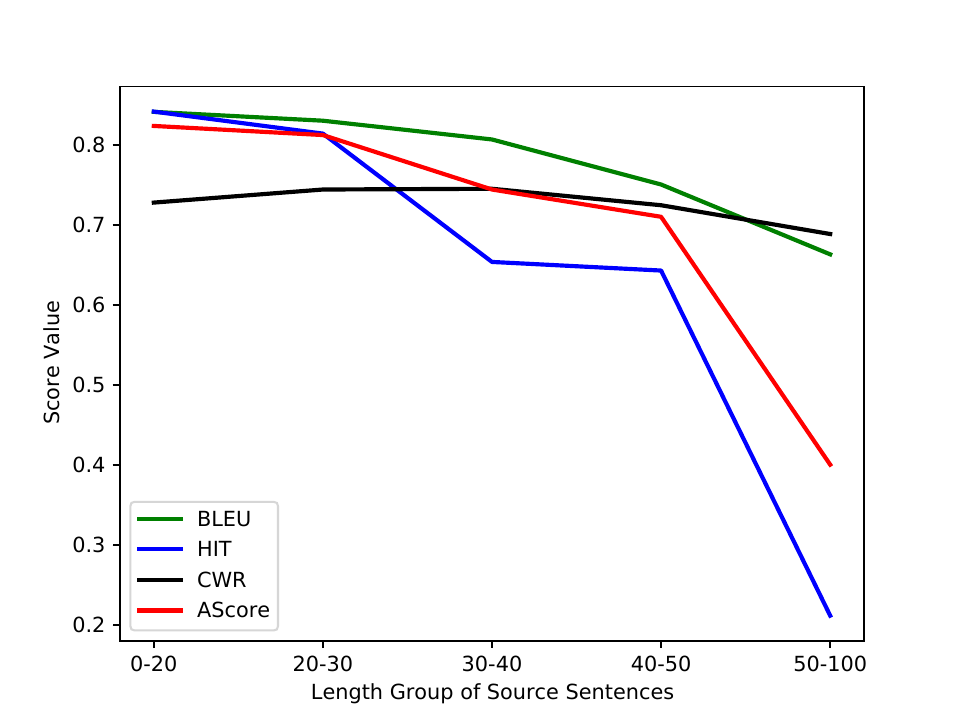}
\vspace{-0.1in}
\caption{Sentence length v.s. performance.}
\label{image:Length}
\end{figure}

\noindent\textbf{Insight Analysis}
To analyze the influence of the sentence length on the model performance, we divide the source sentences into five groups with different length and calculate the average scores among different length groups. The results are shown in Figure~\ref{image:Length}.

We can observe that with the increase of the sentence length, the values of BLEU, HIT, and AScore drop, which is in accord with traditional machine translation tasks. However, the trend of the CWR score is different from that of the other three metrics, which keeps a stable performance. The reason is that CWR reflects a language style feature, which is relatively independent from the length.
From this experiment, we can conclude that a single CWR score can not reveal the actual performance of the model in our task.
These results also confirm the reasonableness of the design of AScore, which assigns more importance weights to the BLEU and HIT scores compared with the CWR score.

\begin{figure}[h]
\centering
\includegraphics[width=0.45\textwidth]{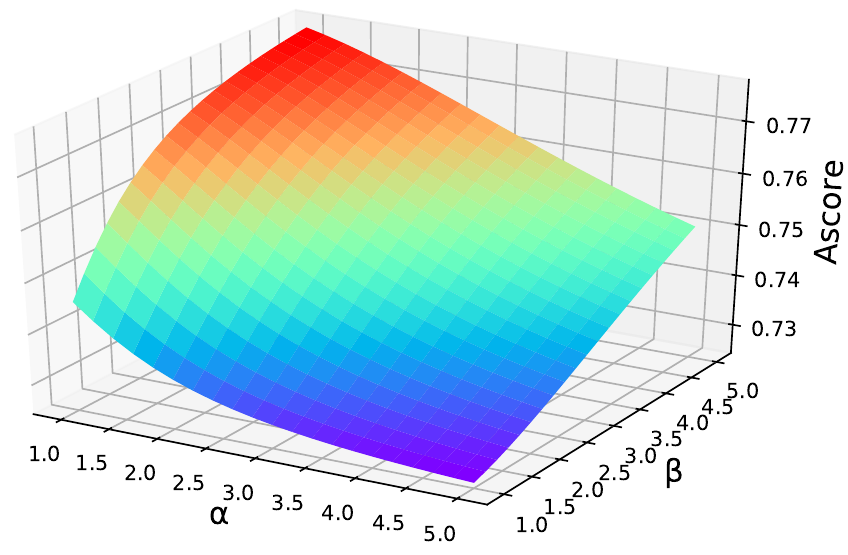}
\caption{Ascore changes regrading $\alpha$ and $\beta$.}
\label{image:parameter}
\end{figure}

\smallskip
\noindent\textbf{Hyperparameter Analysis}.
In the proposed new metric AScore, there are two key parameters $\alpha$ and $\beta$. Figure~\ref{image:parameter} shows the values of AScore with regard to the changes of $\alpha$ and $\beta$. We can observe that given a fixed $\alpha$, the values of AScore increase with the increase of $\beta$. When we fix the value of $\beta$, the values of AScore will decrease with the increase of $\alpha$. The values of AScore are in the range [BLEU, HIT]. Thus, AScore is a trade-off between readability (BLEU), correctness (HIT), and simplicity (CWR). 

\begin{table*}[!htbp]
\centering
\resizebox{\textwidth}{!}{
\begin{tabular}{|p{0.15\linewidth} | p{0.8\linewidth}|}
\hline
Source:      & {\underline{\textbf{NSTEMI}}/\underline{\textbf{CAD}} - history of \underline{\textbf{3V-CABG}} with only \underline{\textbf{RCA}} graft still patent .}\\ \hline
Reference 1:     & {\textcolor{red}{{[}non-ST-elevation myocardial infarction{]}}/\textcolor{red}{{[}coronary artery disease{]}} - history of \textcolor{red}{{[}coronary artery bypass graft{]}} with only \textcolor{red}{{[}right coronary artery{]}} graft still patent .}  \\ \hline
Reference  2:     & {\textcolor{blue}{heart attack}/\textcolor{blue}{heart disease} - history of \textcolor{blue}{heart bypass surgery} with only \textcolor{blue}{right heart artery} graft still patent .}  \\ \hline\hline
{\namemodel} & {\textcolor{blue}{heart attack attack}/\textcolor{blue}{heart disease}-history of \textcolor{red}{coronary artery bypass graft} with only \textcolor{blue}{right heart artery} graft still patent .}\\ \hline
BERT-MT & {\textcolor{blue}{heart attack/heart disease} - history of 3v - \textcolor{blue}{heart bypass surgery} with only \textcolor{blue}{right right heart artery} graft still patent . }\\ \hline
EditNTS & {nstemi/cad - history of 3v-cabg with only right heart artery still patent . eost}\\ \hline
T5 & {NSTEMI/CAD-history of 3V-CAD with only RCA graft still patent}\\ \hline
BART & {NSTEMI/\textcolor{red}{coronary artery disease}-history of 3V-catheter graft with only \textcolor{red}{right coronary artery} graft still patent}\\ \hline
Seq2Seq      & {- history of with only right heart artery graft . } \\ \hline
Seq2Seq$-$ & {- - history of with history only - when are only . }\\ \hline
PointerNet      & {- history of right heart disease graft with two-vessel coronary artery still patent . }\\ \hline
Moses        & {nstemi/cad - history of 3v-cabg with only still patent artery graft . } \\ \hline
\end{tabular}
}
\caption{An example that {\namemodel} outperforms other baselines.}
\label{Table:expsen}
\end{table*}

\begin{table*}[!htb]
\centering
\resizebox{\textwidth}{!}{
\begin{tabular}{|p{0.15\linewidth} | p{0.8\linewidth}|}
\hline
Source:      & {\# \underline{\textbf{cirrhosis}} : patient with history of alcoholic vs \underline{\textbf{nash}} \underline{\textbf{cirrhosis}} complicated by esophagel , gastric , and rectal varices } \\ \hline
Reference 1:     & {\# \textcolor{red}{[chronic disease of the liver]} : patient with history of alcoholic vs \textcolor{red}{[non-alcoholic steatohepatitis]} \textcolor{red}{[chronic disease of the liver]} complicated by esophagel , gastric , and rectal varices .} \\ \hline
Reference 2:     & {\# \textcolor{red}{[chronic disease of the liver]} : patient with history of alcoholic vs \textcolor{blue}{liver inflammation and damage} complicated by esophagel , gastric , and rectal varices .}\\ \hline\hline
{\namemodel} & {cirrhosis : patient with history of alcoholic vs \textcolor{red}{[non-alcoholic]} cirrhosis cirrhosis cirrhosis complicated by esophagel , gastric , and rectal varices .} \\ \hline
BERT-MT & {cirrhosis cirrhosis : patient with history of alcoholic vs late disease disease complicated by food , , , , , , , eseseseseseseseseseseseseses . } \\ \hline
EditNTS & {\# cirrhosis : patient with history of alcoholic vs nash cirrhosis complicated by esophagel , gastric , and rectal varices . eost} \\ \hline
T5 & {\#cirrhosis: patient with history of alcoholic vsnash cirrhosis complicated by inflammation of the esophagus, gastric, and rectal varices.}\\ \hline
BART & {\#cirrhosis: patient with history of alcoholic vsnash cirrhosis complicated by esophagel, gastric, and rectal varices}\\ \hline
Seq2Seq      & {\# cirrhosis : patient with history of alcoholic vs cirrhosis cirrhosis , gastric , gastric , and , , , varices .} \\ \hline
Seq2Seq$-$ & {\# surgical history of patient with history by , history , patient , by surgical by and by surgical tract . } \\ \hline
PointerNet      & {\# cirrhosis : patient with history of painful cell function cirrhosis complicated by , , , , , , and rectal rectal in rectal varices .}\\ \hline
Moses        & {\# cirrhosis : patient with history of alcoholic cirrhosis , complicated by nash esophagel , acid , and rectal and . }  \\ \hline
\end{tabular}
}
\caption{A hard example that all the approaches cannot translate accurately.}
\label{Table:expsenhard}
\end{table*}

\smallskip
\noindent\textbf{Case Study}
To further demonstrate the effectiveness of the proposed {\namemodel}, we conduct a case study as shown in Table~\ref{Table:expsen}. We can observe that {\namemodel} successfully and correctly translates all the professional terms. The BLEU score of {\namemodel} is 0.8235, which is higher than those of other baselines. 

{\namemodel} first uses the locator to identify the professional terms ``NSTEMI'', ``CAD'', ``3V-CABG'', and ``RCA'', which are hard to be understood by patients with low health literacy. Then the neural interpreter selects the best replacement from the mapping dictionary for each professional clinical term, such as replacing ``CAD'' with ``coronary artery disease''. Finally, the polisher is in charge of simplifying the clinical language to layperson-understandable languages, such as translating ``coronary artery disease'' to ``heart disease''. BERT-MT also generates high-quality sentences, but there are missed professional and redundant words, such as ``3v - '' and ``right''. Thus, the readability of BERT-MT's output is lower than that of {\namemodel}'s.

Table~\ref{Table:expsenhard} shows a hard example that almost all the approaches fail to translate the source sentence. Compared with other baselines, {\namemodel} can generate the word ``non-alcoholic'', which leads to its performance better than others. However, all the approaches are unfamiliar with the word ``cirrhosis'', which is the main reason for the failure. From these results, we can find that it is challenging to accurately translate clinical jargon to layperson-understandable language.

In both cases, the EditNTS failed to simplify any professional words. This result can prove that despite the general ATS methods are good at simplifying complex general words and sentences, they are limited in their simplification of professional medical terminologies.
    

\section{FKGL Score Results}\label{appedix:fkgl}
\begin{table*}[!htb]
\centering
\resizebox{\textwidth}{!}{
\begin{tabular}{|c|c|c|c|c|c|c|c|c|c|c|c|c||c|}
\hline
\textbf{Model}            & \textbf{BLEU-1}                         & \textbf{BLEU-2}  & \textbf{BLEU-3}  & \textbf{BLEU-4}  & \textbf{BLEU} & \textbf{METEOR} & \textbf{ROUGE-L} & \textbf{CIDEr} & \textbf{SARI} & \textbf{HIT}    & \textbf{CWR}    & \textbf{AScore} & \textbf{FKGL$\downarrow$}\\ \hline
GroundTruth & -                       & - & - & - & - & - &- & - & - & - & - & - & 10.9603\\ \hline
Dictionary & 0.7158                       & 0.6364 & 0.5684 & 0.5076 & 0.6070 & 0.3933 & 0.7308 & 4.2037 & 37.3391 & 0.5572 & 0.6407 & 0.5948 & 10.3001 \\ \hline
Moses      & 0.7880                       & 0.7130 & 0.6530 & 0.6016 & 0.6889 & 0.4237 & 0.8188 & 5.1046 & 51.6827 & 0.6823 & 0.7543 & 0.6859 & 9.0255\\ \hline
Seq2seq    & 0.7136                       & 0.6322 & 0.5969 & 0.5160 & 0.6147 & 0.3533 & 0.7609 & 4.1299 & 46.1328 & 0.7388 & 0.7980 & 0.6648 & 8.1309\\ \hline
Seq2seq-   & 0.5066                       & 0.3315 & 0.2373 & 0.1787 & 0.3135 & 0.1859 & 0.4948 & 1.2670 & 24.5346 & 0.6427 & \textbf{0.8367} & 0.4070 & 6.4085\\ \hline
Seq2seq-S  & 0.7180                       & 0.6386 & 0.5778 & 0.5267 & 0.6153 & 0.3604 & 0.7683 & 4.2635 & 46.5085 & 0.7331 & 0.7953 & 0.6630 & 8.8005\\ \hline
PointerNet & 0.6870                       & 0.5904 & 0.5158 & 0.4541 & 0.5618 & 0.3338 & 0.7285 & 3.9458 & 42.2857 & 0.6414 & 0.7555 & 0.5949 & 9.4993\\ \hline
EditNTS    & 0.8213                       & 0.7801 & 0.7452 & 0.7132 & 0.7649 & 0.4674 & 0.7401 & 5.9508 & 62.6036 & 0.6405 & 0.6915 & 0.7116 & 5.4448\\ \hline
BART       & 0.7148                       & 0.6755 & 0.6396 & 0.6060 & 0.6590 & 0.5320 & 0.7616 & 4.9783 & 70.3058 & 0.5266 & 0.7311 & 0.6191 & 10.5039\\ \hline
T5         & 0.7223                       & 0.6812 & 0.6445 & 0.6103 & 0.6646 & 0.5305 & 0.7645 & 5.0629 & 71.3255 & 0.5262 & 0.7342 & 0.6220 & 10.7484\\ \hline
BERT-MT    & 0.8003                       & 0.7428 & 0.6952 & 0.6531 & 0.7228 & 0.4566 & 0.8218 & 5.3293 & \textbf{72.2260} & 0.7808 & 0.7358 & 0.7417 & 9.0255\\ \hline
 \rowcolor{lavendermist} {\namemodel} & \textbf{0.8624}               & \textbf{0.8291} & \textbf{0.8004} & \textbf{0.7737} & \textbf{0.8165} & \textbf{0.5290} & \textbf{0.8894} & \textbf{6.7212} & 70.8583 & \textbf{0.7986} & 0.7328 & \textbf{0.7983} & 10.3187\\ \hline
 \rowcolor{lavendermist}$\uparrow$    & +7.8\% & +11.6\% & +15.7\% & +18.5\% & +12.9\% & +15.9\% & +8.2\% & +26.1\% & -1.9\% & +2.2\% & -12.4\% & +7.6\% & \\\hline
\end{tabular}
}
\vspace{-0.1in}
\caption{Results with FKGL score.}
\label{Table:expwithfkgl}
\vspace{-0.2in}
\end{table*}
For the FKGL (lower the better), we can find that the ground truth is the worst result as shown in Table~\ref{Table:expwithfkgl}, which can effectively illustrate why we argue that it is not a suitable metric for our task. The simplification of the professional medical terms involves replacing the short abbreviations into long common words. However this will actually increase the FKGL score since FKGL focus on the average words in a sentence and the average syllables in a word. Transferring the abbreviations into long common words will increase the above metrics and results a worse FKGL score.
\end{document}